\documentclass[a4paper,twoside]{article}

\usepackage{epsfig}
\usepackage{subcaption}
\usepackage{calc}
\usepackage{amssymb}
\usepackage{amstext}
\usepackage{amsmath}
\usepackage{amsthm}
\usepackage{multicol}
\usepackage{pslatex}
\usepackage{apalike}
\usepackage[bottom]{footmisc}
\usepackage{SCITEPRESS}     

\usepackage{graphicx}
\usepackage{multirow}
\usepackage{booktabs}
\usepackage{multicol}
\usepackage{url}     
\usepackage{float}
\usepackage{rotating} 

\usepackage{pythonhighlight}

\usepackage{listings}
\usepackage{subcaption}

\usepackage{caption}
\DeclareCaptionType{pcode}[Pseudo-Code][List of Pseudo-Code]

\usepackage{tcolorbox}
\usepackage[scaled=.95]{inconsolata}

\begin{document}

\title{Induced Feature Selection by Structured Pruning}

\author{
\authorname{{Nathan Hubens $^{\star \dagger}$ \thanks{This research has been conducted in the context of a joint-PhD between the two institutions.} \qquad {Victor Delvigne} $^{\star \ddagger}$ \qquad Matei Mancas $^{\star}$ \\  Bernard Gosselin $^{\star}$  \qquad  Marius Preda $^{\dagger}$  \qquad Titus Zaharia $^{\dagger}$ }}
\affiliation{$^{\star}$ ISIA Lab, University of Mons, Belgium \\
      $^{\dagger}$ Artemis, IP Paris, France
      \\
      $^{\ddagger}$ CRIStAL, IMT Lille Douai, France}
\vspace{2cm}}



\abstract{The advent of sparsity inducing techniques in neural networks has been of a great help in the last few years. Indeed, those methods allowed to find lighter and faster networks, able to perform more efficiently in resource-constrained environment such as mobile devices or highly requested servers. Such a sparsity is generally imposed on the weights of neural networks, reducing the footprint of the architecture. In this work, we go one step further by imposing sparsity jointly on the weights and on the input data. This can be achieved following a three-step process: 1) impose a certain structured sparsity on the weights of the network; 2) track back input features corresponding to zeroed blocks of weight; 3) remove useless weights and input features and retrain the network. Performing pruning both on the network and on input data not only allows for extreme reduction in terms of parameters and operations but can also serve as an interpretation process. Indeed, with the help of data pruning, we now have information about which input feature is useful for the network to keep its performance. Experiments conducted on a variety of architectures and datasets: MLP validated on MNIST, CIFAR10/100 and ConvNets (VGG16 and ResNet18), validated on CIFAR10/100 and CALTECH101 respectively, show that it is possible to achieve additional gains in terms of total parameters and in FLOPs by performing pruning on input data, while also increasing accuracy.}

\onecolumn \maketitle \normalsize \setcounter{footnote}{0} \vfill

\section{\uppercase{Introduction}}
\label{sec:introduction}

Over the last few years, neural networks have become ubiquitous in most AI-based tasks. They are now considered as state-of-the-art techniques in many field but have at the same time followed a Moore-like trend, becoming increasingly deeper, involving more and more parameters. Their high performance is often attributed to their notable over-parameterization, allowing them to have enough expressivity for real-world problems. However, this over-parameterization usually comes with a cost at training and inference, making the training and deployment of neural networks in resource-constrained environment challenging. \\

For that reason, many research about efficient neural network design has been conducted recently. In particular, one of the most established technique to reduce neural networks consumption footprint consists in removing useless parameters from a network; the so-called parameter pruning. It has been recently hypothesized that all networks possess a subnetwork that is at least as effective and can be trained at least as fast than the whole network, as long as they start from the same initial conditions \cite{lottery}. Such a subnetwork has to be discovered by performing several rounds of pruning and reinitialization of remaining weights to their original value. Pruning can thus be seen as a compression technique as it allows to reduce the number of parameters in a network and, in some cases, allows it to run faster. But it can also be seen as a regularization method, as it reduces the influence of useless parameters and, for certain amount of sparsity, can help networks to increase their generalization abilities \cite{torsten}. \\

Introducing weight sparsity in a network can be seen as a rigid alternative to dropout regularization technique \cite{dropout} as weights are permanently masked after being zeroed out. Another regularization technique that, this time do not operate by zeroing network's weights like dropout but rather by setting parts of the input to zero, is Cutout \cite{cutout}. This technique removes part of the input image, making the network less dependant to those input pixels and thus regularizing the network. Although a lot of development has been performed on the pruning of networks weights, very few are interested in the permanent pruning of input data. \\


In our work, we show that pruning techniques do not only affect network weight but can, in some cases, affect the input data. By using a weight pruning technique giving us feedback information about input data, it allows us to reduce the size of input data by removing unimportant parts in order to further accelerate network training and inference. This technique thus benefits both from weight and input data pruning for maximal efficiency. \\

Contributions of our work are summarized as following: 
\begin{itemize}
\item Propose a dual weight-input data pruning technique, benefiting both from input data reduction and weight reduction for increased efficiency.
\item Show that the proposed technique allows to obtain indication about important features of the input images.
\item Validate our results on two types of architecture: Multi-Layer Perceptrons (MLP) and Convolutional Neural Networks (ConvNets), each providing a different kind of feedback regarding its input images. 
\end{itemize}

\section{\uppercase{Related Work}}

Pruning techniques have 3 distinct axis of application: the granularity, the criteria and the schedule. \\

\textbf{Granularity. } The pruning granularity concerns decisions about the shape of group of weights that are removed. Granularity is usually separated into two types: 1) unstructured pruning, or when pruning is performed on individual weights in the network \cite{brain_damage,brain_surgeon2}, and 2) structured pruning, or when weights are removed in blocks \cite{Li,cpa,hub}. Because pruning usually leads to sparse weight matrices, requiring dedicated hardware or software to take advantage of the removal of parameters, structured pruning is often performed on filters. Doing so effectively changes the architecture as filters can now be physically removed from the network, creating a dense but smaller architecture, not requiring any dedicated sparse computation capabilities anymore. \\

\textbf{Criteria. } Pruning criteria is probably the largest segment of pruning research. Second-order approximation of the loss \cite{brain_damage,brain_surgeon2}, $l_1$-norm \cite{han}, movement \cite{movement}, Taylor approximation \cite{taylor} are some of the most commonly encountered criteria. Some work even explored $l_0$ regularization \cite{l0} or variational dropout \cite{var_dropout}. However, despite its apparent simplicity, $l_1$-norm is, to this day, the most commonly used criteria, as it happens to be very consistent across datasets and to lead to comparable or better results than other more complex pruning criteria \cite{state}. \\

\textbf{Scheduling. } Pruning schedules concerns how training and pruning are combined. Early pruning methods proposed to remove parameters in a single step from an already trained model, i.e. the so-called one-shot pruning \cite{Li}. Further research showed that performing pruning iteratively, by alternating phases of pruning and fine-tuning was able to provide better results \cite{han,iterative}. More recently, alternatives to prune the network during the training \cite{to_prune,ocp} or even before it started to train \cite{snip} have been proposed, removing the needs for a lengthy process before obtaining the final, pruned network. \\


In this work, we mainly focus on the pruning granularity axis as we want to consider granularities that impact input data. In particular, we investigate on how pruning the weights can lead to a pruning of input data, taking advantage of both a reduction of parameters and a reduction of input data.

\section{\uppercase{Methodology}}
\label{sec:sched}

In this paper, we would like to highlight the impact that some structured pruning techniques can have on the input data and take advantage of it to further accelerate our networks. We consider two types of architecture: MLP and ConvNets, both having very different implications on how the pruning methods are applied. In both cases, we will consider image as input data for those architectures.

\subsection{Pruning of MLPs}

\begin{figure}[!htb]
  \centering
  \begin{subfigure}[t]{.48\linewidth}
    \centering\includegraphics[width=\linewidth]{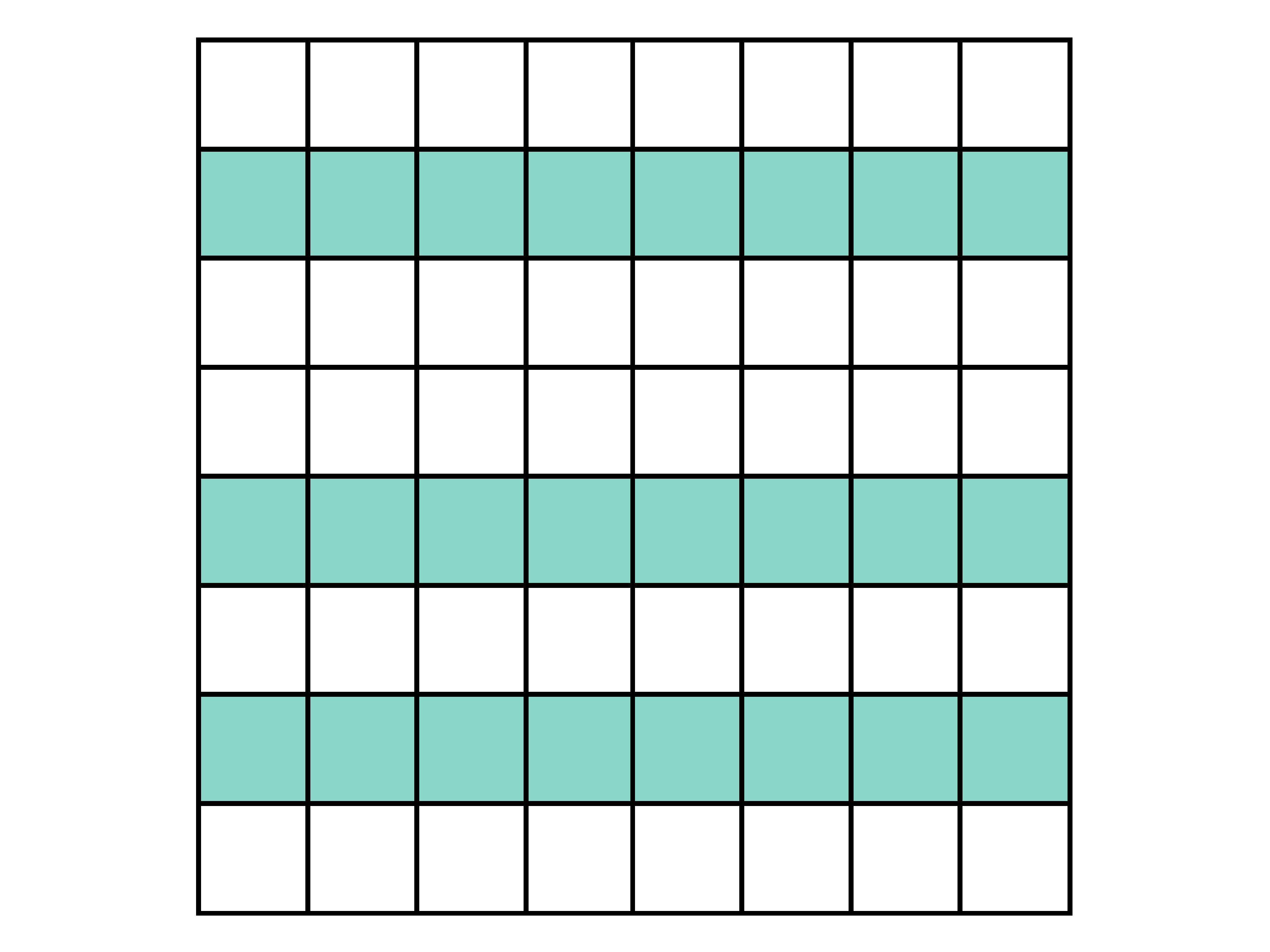}
    \caption{Row Pruning}
  \end{subfigure}
  \begin{subfigure}[t]{.48\linewidth}
    \centering\includegraphics[width=\linewidth]{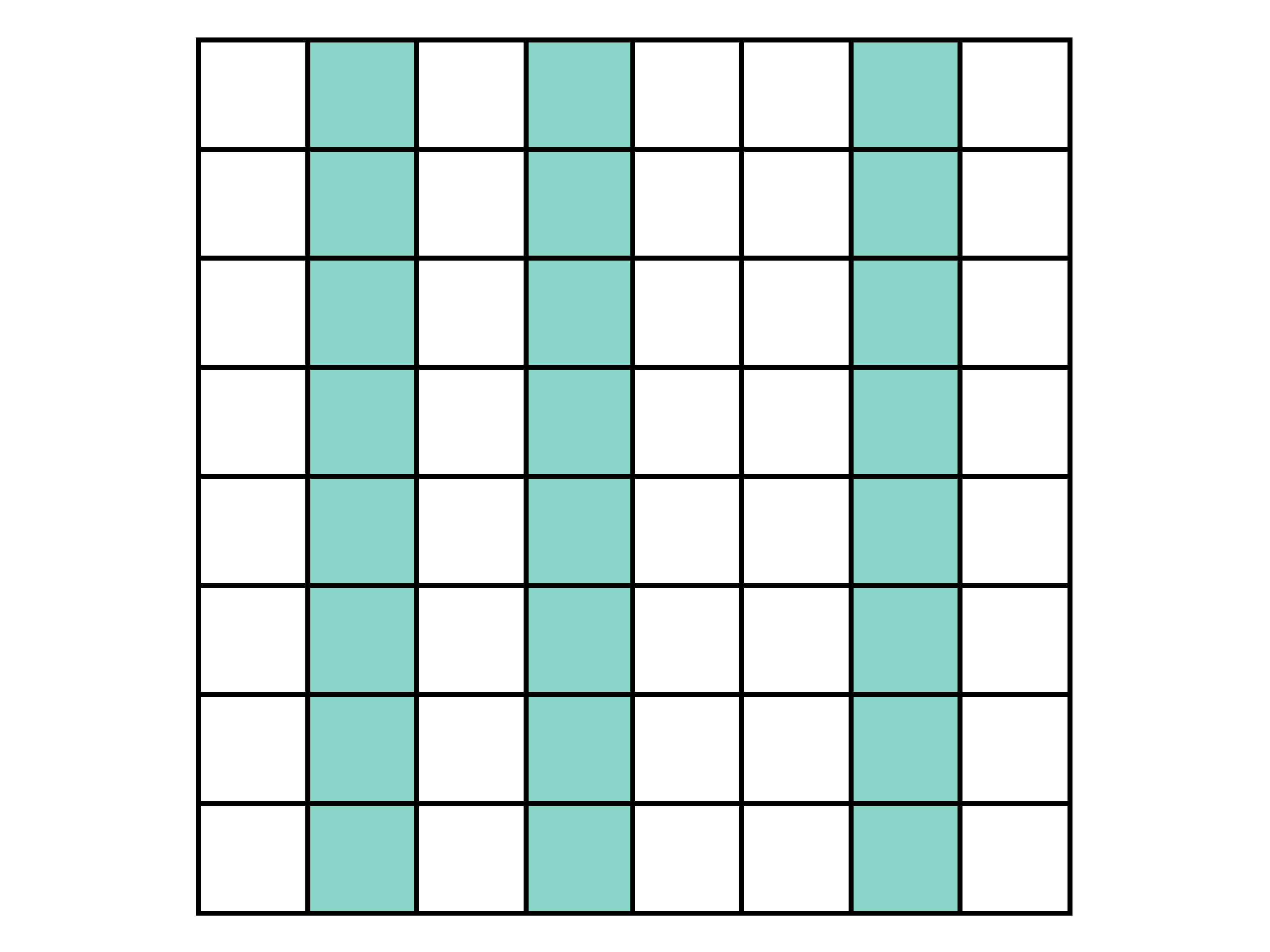}
    \caption{Column Pruning}
  \end{subfigure}
  \medskip
 \caption{Structured sparsity in Linear weight matrices. Zeroed weights are in color.}
 \label{mlp_pruning}
\end{figure}

In MLPs, layers weight are contained in 2D matrices of shape $M\times N$, with $M$ and $N$ being respectively the dimension of input and output. When sparsity is induced in such weight matrices, it is usually performed in an unstructured manner, i.e. on individual weights, leading to sparse weight matrices, difficult to accelerate on common hardware due to poor data locality and low parallelism. However, as represented in Figure \ref{mlp_pruning}, sparsity can also be induced in a structured way, zeroing either complete row or columns of those weight matrices, which makes it easier to accelerate. We express in Pseudo-Code \ref{pc: method_mlp} the different methods of structured weight selection, following \texttt{numpy} slicing standards. \\

\begin{pcode}[!htb]
\begin{python}
Weights = Array(M, N)

Column (1D) = Weights[m, :] 
Row (1D) = Weights[:,n]
\end{python}
\caption[Method selection]{Selection of 1D structures in linear weight matrices.}
\label{pc: method_mlp}
\end{pcode}

\begin{figure}[!htb]
\centering
 \includegraphics[width=0.47\textwidth]{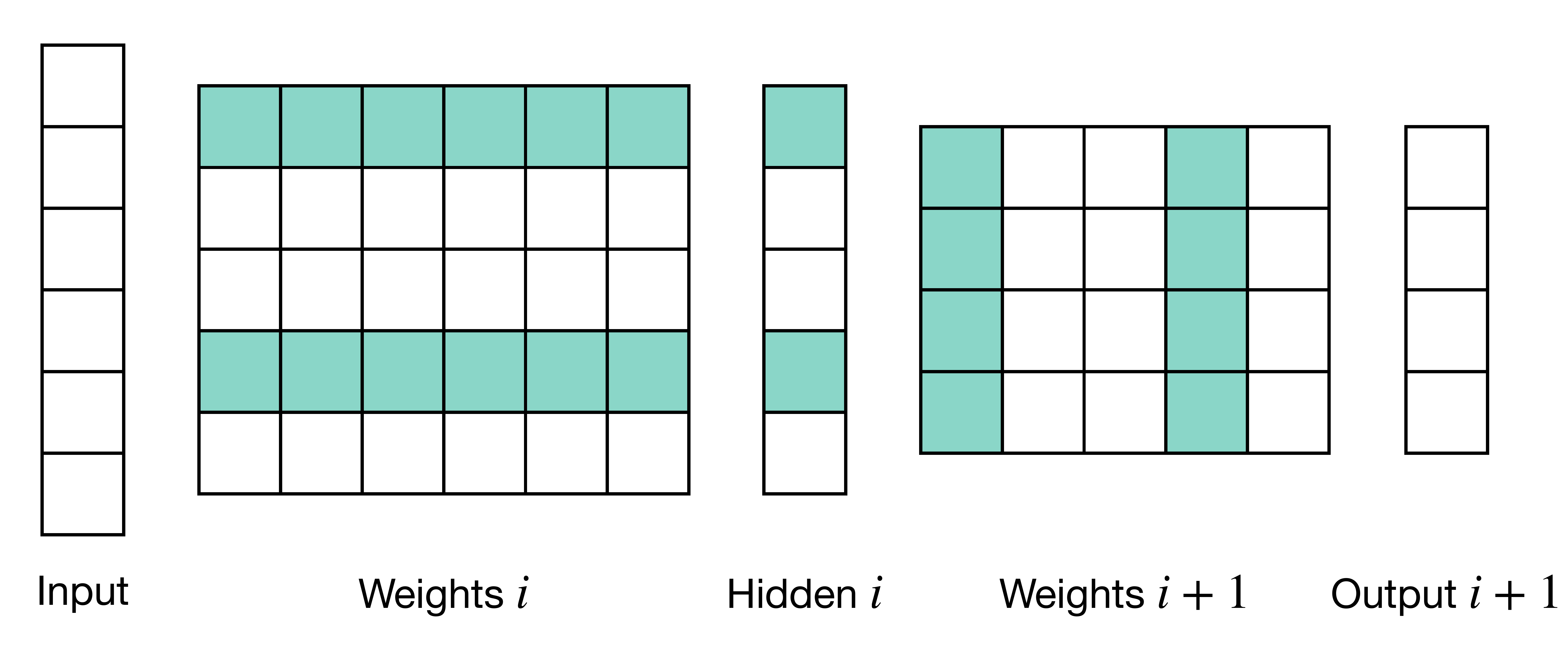}
\caption{Pruning columns in layer $i$ impacts its output and consequently the rows of layer $i+1$. Zeroed weights are in color.}
 \label{mlp_structured_pruning}
\end{figure}

As depicted in Figure \ref{mlp_structured_pruning}, when zeroing a complete row, this has for effect to also zero out the corresponding output result and, as the next layer now has input of smaller dimension, the following weight matrix must be adapted accordingly, i.e. corresponding columns are zeroed out. \\

We might go one step further by actually removing those sparse rows and columns from the initial weight matrices, thus keeping the network dense. Doing so impacts either the inputs or outputs of each weight layer as the weight matrices are now smaller. When selection of which parameters to prune is performed on rows, columns of following matrices are impacted as a side effect. But the opposite is also true; we might base our selection on columns to prune, thus impacting rows in previous weights matrices. \\

In particular, we are interested in the relationship between input data and weight pruning. For that reason, we would like to apply column pruning on the very first layer so that, the input data coming into the network is impacted. Because each column corresponds to an incoming data point, we can track back and remove those data points as they have no longer utility. \\

When inputs are images, a preprocessing step is usually a flattening of the image to a single-dimension vector. For this reason, when removing a single column of the first weight matrix, we can track back to which value in the flattened vector and thus, to which input pixel it corresponds. Removing a single column in the first layer implies the removal of a single pixel in all images from our dataset. We can thus decide on the amount of input information to keep by increasing/decreasing the sparsity in the first weight matrix.

\subsection{Pruning of ConvNets}


There exist many structured pruning techniques for ConvNets as there exists many ways to define blocks of parameters when dealing with 4D weight tensors. For example, we can decide to prune vectors (1D), kernels (2D), slices of weights (2D). However, one of the most efficient ways is to prune complete convolution filters (3D). Indeed, once a filter is completely zeroed out, it can be removed from the network, leading to a dense but smaller architecture and removing the need for specialized hardware or software for acceleration.\\

\begin{figure}[!htb]
\centering
 \includegraphics[width=0.5\textwidth]{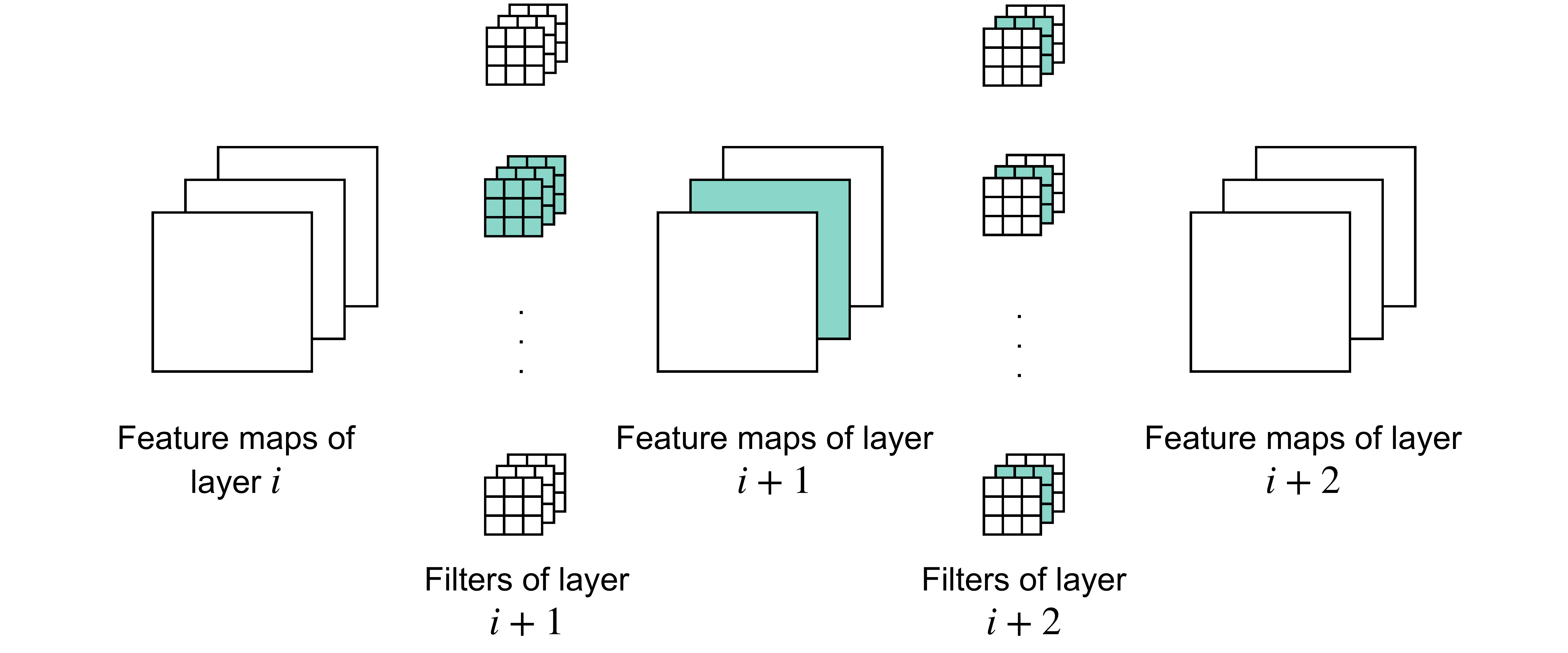}
\caption{The removal of a filter in layer $i+1$ affects the kernels in layer $i+2$, which should also be removed. Zeroed weights are in color.}
 \label{filter_pruning}
\end{figure}

There is one subtlety however as removing the zeroed filter is not enough for the architecture to be operable. When removing a filter, it changes the output shape of the concerned layer as there is one less feature map. This means that the following convolutional layer now receives a smaller input and thus, in all of its filters, the kernel corresponding to the removed feature map has to be removed. As depicted in Figure \ref{filter_pruning}, removing a single filter in layer $i+1$ results in the removal of its corresponding feature maps and of the corresponding kernels in layer $i+2$.\\

As demonstrated, filter pruning is thus the equivalent for ConvNets to row pruning for MLPs as they both impact the output of their layers. But the equivalent to MLPs column pruning for ConvNets is what we call shared-kernel pruning, as all convolution filters have to share the same zeroed kernel in order to impact their input data. Referring to Figure \ref{filter_pruning}, pruning a shared-kernel would mean that we chose to prune the same kernel in each filter of layer $i+2$. As a result, their corresponding input feature map is now useless and can be removed, as well as the corresponding filter in layer $i+1$.\\

\begin{figure}[!htb]
  \centering

  \begin{subfigure}[t]{.48\linewidth}
    \centering\includegraphics[width=1.1\linewidth]{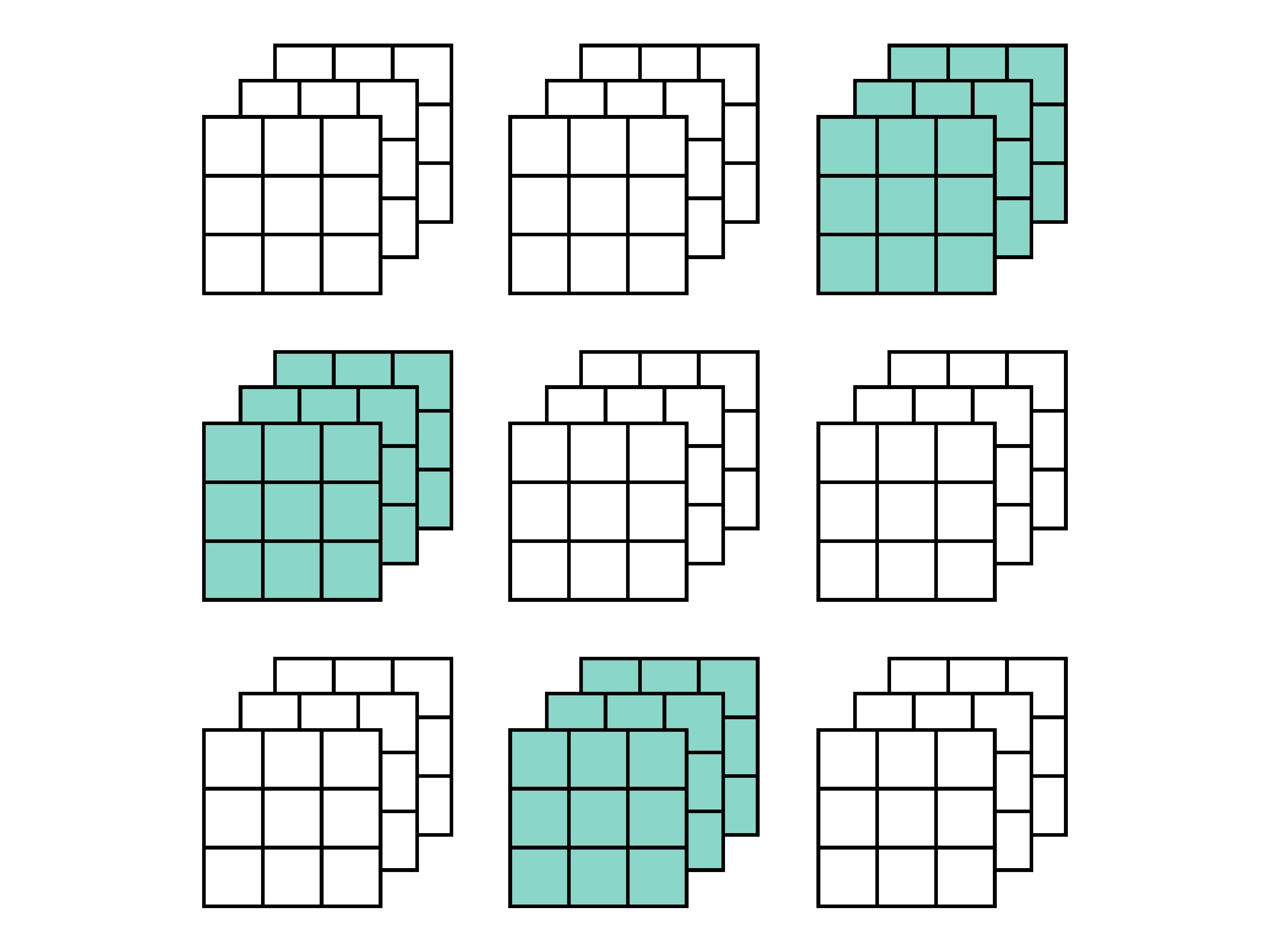}
    \caption{Filter Pruning}
  \end{subfigure}
  \begin{subfigure}[t]{.48\linewidth}
    \centering\includegraphics[width=1.1\linewidth]{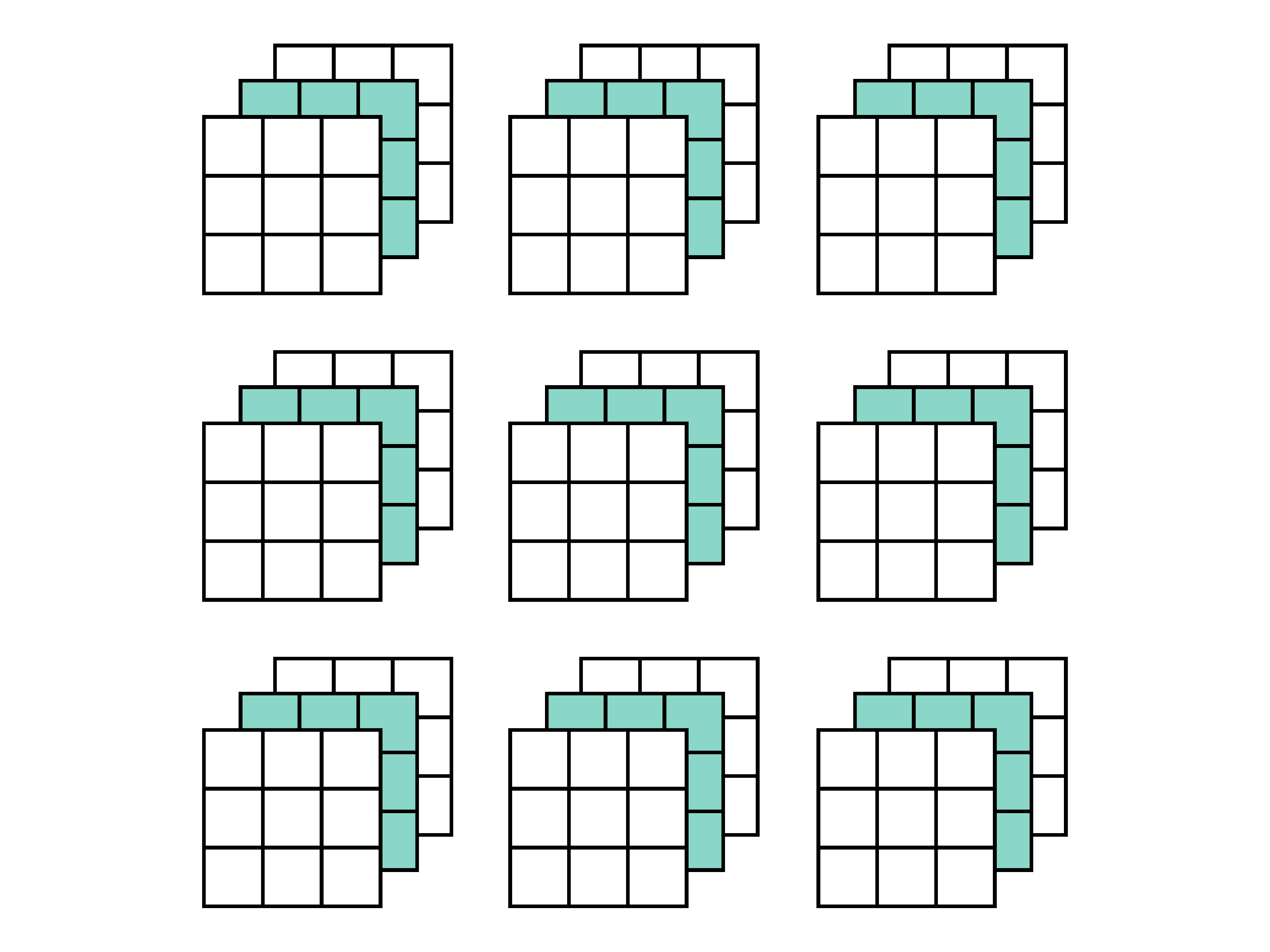}
    \caption{Shared-Kernel Pruning}
  \end{subfigure}
  \medskip
 \caption{Structured granularities for ConvNets.}
 \label{fig:structured_cnn}
\end{figure}

Both technique, represented in Figure \ref{fig:structured_cnn} and whose selection is given by Pseudo-Code \ref{pc: method_cnn}, can thus be seen as two sides of the same coin, removing filter impacting the output of each layer and removing shared-kernels impacting their input. However, those two approaches have a key difference at the very first and very last convolutional layers. Indeed, in the case of shared-kernel granularity on the very first layer, the previous inputs are the whole model inputs. Removing a shared-kernel will thus be equivalent to removing a channel to all input images. \\

This last is the effect that we want to take advantage of, i.e. shifting the decision about weights to remove by selecting shared-kernels, effectively reducing the input dimension, as well as the network size for further size and speed gains. \\

\begin{pcode}[!htb]
\begin{python}
Weights = Array(O, I, Kx, Ky)

Shared-Kernel (3-Dim) = Weights[:,i,:,:]
Filter(3-Dim) = Weights[o, :, :, :] 
\end{python}
\caption[Method selection]{Selection of 3D structures in convolutional weight matrices. With $O$, $I$, $K_x$, $K_y$ being respectively the output and input number of channels and the horizontal and vertical dimensions of the kernels.}
\label{pc: method_cnn}
\end{pcode}

However, because ConvNets hold more inductive biases than MLPs, especially the weight sharing that allows ConvNets to reuse the same weights filters for the complete input image, information about individual pixels is lost. For this reason, the only information feed-back that can be obtained from structured pruning is about which channel to keep. \\



In this paper, we propose to apply this shared-kernel pruning to our ConvNet architectures, allowing the pruning process to remove parts of the input image that are judged as less useful, thus giving us feedback about our input data.

\section{\uppercase{Experiments}}

In this paper, we conduct 3 types of experiments:

\begin{itemize}
    \item \textbf{Prune Weights}: this step consists in zeroing weights considered to be useless according to a chosen metric. Pruning granularities are chosen to be structured and impacting the input dimension, whether it is for MLP (column pruning) or ConvNets (shared-kernel pruning).
    \item \textbf{Prune Data}: based on weights that have been set to zero in previous experiment, we can deduce important parts of the input data that should be retained (either pixels for MLP or channels for ConvNets). To study the real importance of those parts, we retrain a complete dense model, but with the adjusted pruned input data.
    \item \textbf{Prune Both}: in this last step, we remove zeroed out blocks of weights from our sparse model, as well as the corresponding parts of our input data, and retrain the remaining together, giving as much parameter reduction and speed-up as possible to the resulting model. \\
\end{itemize}


\textbf{Pruning Methods.} We apply structured pruning on granularities chosen to affect input dimensions of each layer. This corresponds to removing shared-kernels in the case of ConvNets and columns of weights in the case of MLP. The criteria according to which weight importance is evaluated is the $l_1$-norm, i.e. weights having the lowest are considered as less useful. The same sparsity is applied equally to all network layers (local pruning) and the schedule followed for pruning is the One-Cycle Pruning, gradually adjusting the sparsity level during the training \cite{ocp}. \\

\textbf{Datasets and Architectures.} For our experiments, the datasets have been chosen to be various in terms of image resolution and number of classes. In particular, for our MLP experiment, the network is a simple 3-Layer MLP, trained on three datasets: MNIST \cite{mnist}, CIFAR10, CIFAR100 \cite{cifar}. We also evaluate our method on ConvNet architectures, namely VGG16 \cite{vgg} and ResNet18 \cite{rn18}, also trained on three datasets: CIFAR10, CIFAR100 \cite{cifar} and CALTECH101 \cite{caltech}. \\

\textbf{Training Procedure.} The networks we use for our experiments are trained from randomly initialized weights. When experimenting with the MLP, input images do not undergo any augmentation. When experimenting with ConvNets however, images are first resized to $224\times 224$ and are augmented by using horizontal flips, rotations, warping and random cropping. MLP models are trained for $20$ epochs, ConvNets are for $50$ epochs, with a learning rate warmup method \cite{1cycle} until a nominal value of $0.001$, then gradually decay until the end of the training. \\

\textbf{Frameworks and Hardware. } Experiments are conducted using the PyTorch \cite{pytorch} and fastai \cite{fastai} libraries for the implementation of the neural networks and the training loop, and fasterai \cite{fasterai} for the implementation of the pruning methods. A 12GB Nvidia GeForce GTX 1080 Ti GPU is used for computation.

\section{\uppercase{Results}}

We describe in this section the results obtained for the considered types of architecture: MLP and ConvNets.

\subsection{MLP}

The chosen architecture is a 3-layer MLP, with ReLU non-linearities between each computation layer. For datasets involving color images, we first convert them to grayscale, following a weighted average of color channels \cite{color}, to keep the same architecture for all MLP experiments. The first experiment conducted is thus a classic structured pruning, zeroing out columns in each weight layer according to the desired sparsity. The second experiment takes feedback from the zeroed columns of the first layer, removing the corresponding pixels for each input images and training a dense network with remaining pixels. Finally, the last experiment consists in combining information gathered from the two previous experiments, i.e. remove sparse columns of the pruned network and retrain it with pixels remaining from previous experiment. \\

We report in Table \ref{table:vgg16} the results for 3 levels of sparsity, namely $25\%$, $50\%$ and $75\%$. Overall, we observe a similar trend for all considered datasets. Indeed, all of them perform better when pruning is applied both on data and on weights. In the case of CIFAR10 and CIFAR100, it is important to mention that pruning data seems to help the network to generalize better.

\begin{table}[!htbp]
\begin{center}
\resizebox{0.45\textwidth}{!}{
\renewcommand{\arraystretch}{1.1}
\begin{tabular}{l|l||c c c}
\cmidrule[1pt]{3-5} 
\multicolumn{2}{c}{}&\multicolumn{1}{c}{Prune Weights }&\multicolumn{1}{c}{Prune Data }&\multicolumn{1}{c}{Prune Both}\\\midrule
\multicolumn{2}{c}{\textbf{MNIST}} & \multicolumn{2}{c}{}             &  \\\midrule
\multirow{3}{4mm}{\begin{sideways}\parbox{12mm}{Sparsity}\end{sideways}}
& 25\% &  97.79 $\pm$ 0.12 & 97.71 $\pm$ 0.16  &  97.92 $\pm$ 0.14 \\
& 50\% &  97.75 $\pm$ 0.10  &   97.90 $\pm$ 0.13  &  97.97  $\pm$ 0.09 \\
& 75\% &   96.81 $\pm$ 0.20  &   97.30 $\pm$ 0.16  &  97.26 $\pm$ 0.03 \\\midrule

\multicolumn{2}{c}{\textbf{CIFAR10}} & \multicolumn{2}{c}{}             &  \\\midrule
\multirow{3}{4mm}{\begin{sideways}\parbox{12mm}{Sparsity}\end{sideways}}
& 25\% & 40.93 $\pm$ 0.39 &  41.73 $\pm$ 0.21 &  41.06 $\pm$ 0.41  \\
& 50\% &  39.72 $\pm$ 0.37 &  42.05 $\pm$ 0.80  &  41.47  $\pm$  0.61 \\
& 75\% & 38.11 $\pm$ 0.27  &  42.63 $\pm$ 0.33  & 40.69 $\pm$ 0.70 \\\midrule

\multicolumn{2}{c}{\textbf{CIFAR100}} & \multicolumn{2}{c}{}             &  \\\midrule
\multirow{3}{4mm}{\begin{sideways}\parbox{12mm}{Sparsity}\end{sideways}}
& 25\% & 13.71 $\pm$ 0.33 & 14.11$\pm$ 0.19 & 13.15 $\pm$  0.51 \\
& 50\% & 12.90 $\pm$ 0.22 &  14.09 $\pm$ 0.15  &  12.28 $\pm$  0.26 \\
& 75\% &  11.62 $\pm$ 0.14  &  15.01 $\pm$ 0.34  &  10.34 $\pm$  0.76 \\\midrule

   \end{tabular}}
\end{center}
\caption{Results of MLP. Reported values are mean and standard deviation over 5 iterations.}
\label{table:mlp}
\end{table}

\begin{figure}[!htb]
\centering
 \includegraphics[width=0.47\textwidth]{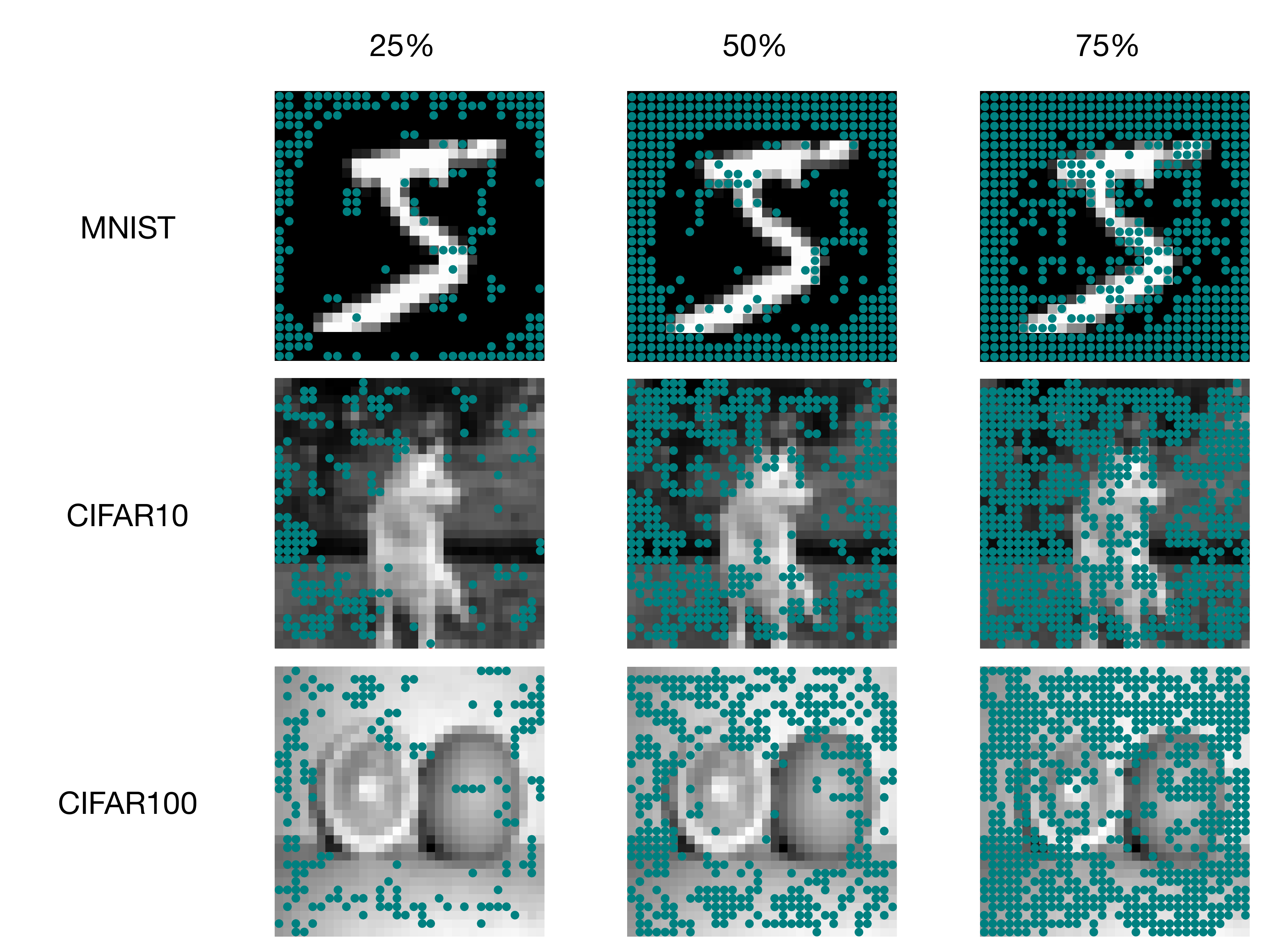}
\caption{Masks of important pixel obtained after performing the pruning weight. Pixels in blue can be removed from all input images to perform the pruning data experiment.}
 \label{mask_mlp}
\end{figure}

By analyzing the removed weights for different sparsity levels, we can retrieve important pixels in our input images. We depict the remaining pixels in Figure \ref{mask_mlp}. We can observe a general trend of preserving central pixels, corresponding to the region where most of important information is contained. \\

\begin{figure}[!htb]
\centering
 \includegraphics[width=0.465\textwidth]{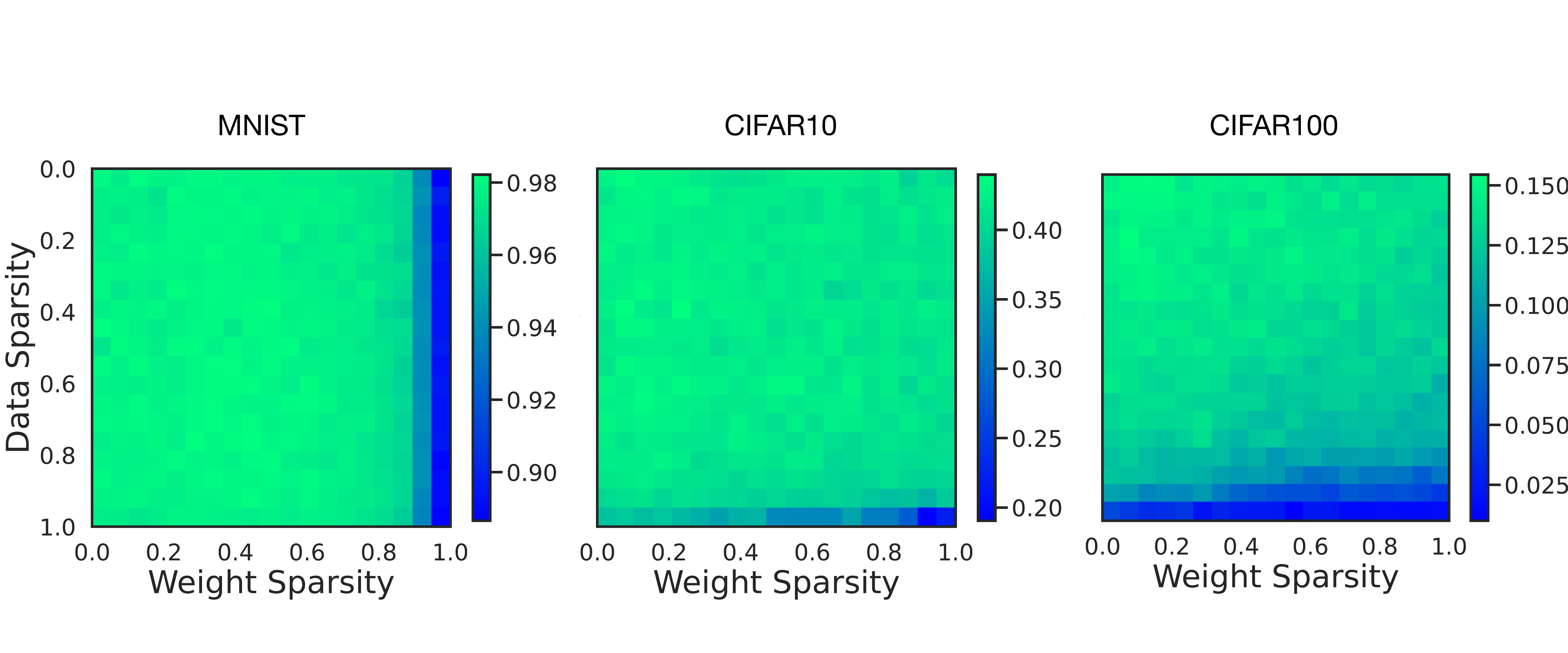}
\caption{Evolution of accuracy when weight sparsity and data sparsity are evolving.}
 \label{mlp_pruning_both}
\end{figure}

We also report results of the mixing of different sparsity levels for data and weights in Figure \ref{mlp_pruning_both}. As can be observed, for MNIST, the final accuracy is predominantly affected by the weight sparsity, while it seems to be invariant to data sparsity. This can be attributed to the vast majority of MNIST pixels being zeroes, thus requiring a high level of data sparsity before reaching important information pixels. On the other hand, CIFAR10 and CIFAR100 are primarily affected by data sparsity, which can be attributed to sparse data forcing the model to focus on important parts of the image, consequently helping it in its learning process.

In Table \ref{tab:budget_MLP}, we report the parameter and FLOPs gains, computed following the Shrinkbench methods \cite{BlalockOFG20}, for a pruning sparsity of $50\%$. In particular, pruning both weights and data allows for a reduction of almost $75\%$ in both parameter number and FLOPs, while providing better accuracy than regular weight pruning. 

\begin{table}[h!]
  \begin{center}
    \setlength\tabcolsep{4pt}
    \small
    \renewcommand{\arraystretch}{1.1}
    \begin{tabular}{l|cc}
    \cmidrule[1pt]{2-3}  
      & \multicolumn{1}{c}{Parameters (k)} & \multicolumn{1}{c}{FLOPs (k)} \\
    \hline
  {{\textbf{Baseline}}} & $105.21$ & $105.00$ \\
  
 {{\textbf{Prune Weights}}} & $52.71  \ (\downarrow 49.90\%)$ & $ 52.50  \ (\downarrow 50.00\%) $ \\
      
  {{\textbf{Prune Data}}} & $ 58.17  \ (\downarrow 44.71\%) $ & $ 57.96  \ (\downarrow 44.80\%) $ \\
      
  {{\textbf{Prune Both}}} & $ 26.57  \ (\downarrow 74.75\%)$ & $26.46 \ (\downarrow 74.80\%)$  \\
      \bottomrule
    \end{tabular}
  \end{center}    
    \caption{Parameter and FLOPs gains for MLP.}
    \label{tab:budget_MLP}
\end{table}

\subsection{Convolutional Models}

We conduct the same three experiments for both of our ConvNets and report results in Table \ref{table:vgg16} for VGG16 and Table \ref{table:resnet18} for ResNet18. For both architectures, we can observe a similar trend in the results, where pruning both our input data and network weights seems to provide better results than weight pruning only. As data pruning is performed on the level of input channels, there are only three valid sparsity levels for data, i.e. removing 0, 1 or 2 channels, leading to respectively $0\%$, $33\%$ and $66\%$ of data sparsity. We thus keep sparsity levels to $25\%$, $50\%$ and $75\%$ in the Table for consistency across experiments but actually apply the nearest smaller sparsity to data, e.g. 1 
channel is removed for $50\%$ sparsity.

\begin{table}[!htbp]
\begin{center}
\resizebox{0.45\textwidth}{!}{
\renewcommand{\arraystretch}{1.1}
\begin{tabular}{l|l||c c c}
\cmidrule[1pt]{3-5} 
\multicolumn{2}{c}{}&\multicolumn{1}{c}{Prune Weights }&\multicolumn{1}{c}{Prune Data* }&\multicolumn{1}{c}{Prune Both}\\\midrule
\multicolumn{2}{c}{\textbf{CIFAR10}} & \multicolumn{2}{c}{}             &  \\\midrule
\multirow{3}{4mm}{\begin{sideways}\parbox{12mm}{Sparsity}\end{sideways}}
& 25\% &  92.97 $\pm$ 0.06  &   94.12 $\pm$ 0.17  & 92.89  $\pm$ 0.21  \\
& 50\% &  92.53 $\pm$ 0.24 &  93.84 $\pm$ 0.22  &   92.68 $\pm$ 0.22   \\
& 75\% &   90.15 $\pm$ 0.31  &  92.52 $\pm$ 0.32  &  90.55 $\pm$ 0.13   \\\midrule

\multicolumn{2}{c}{\textbf{CIFAR100}} & \multicolumn{2}{c}{}             &  \\\midrule
\multirow{3}{4mm}{\begin{sideways}\parbox{12mm}{Sparsity}\end{sideways}}
& 25\% &  74.13 $\pm$ 0.32  &  74.52 $\pm$ 0.28 &  73.89 $\pm$ 0.54  \\
& 50\% &  63.17 $\pm$ 0.41  &  71.87 $\pm$ 0.14  &  63.74 $\pm$  0.71 \\
& 75\% &  62.89 $\pm$ 0.56  &  66.69 $\pm$ 0.12  & 62.33 $\pm$ 0.21 \\\midrule

\multicolumn{2}{c}{\textbf{Caltech101}} & \multicolumn{2}{c}{}             &  \\\midrule
\multirow{3}{4mm}{\begin{sideways}\parbox{12mm}{Sparsity}\end{sideways}}
& 25\% &  76.64 $\pm$ 0.74  &  77.84 $\pm$ 0.09 &  76.82 $\pm$ 0.57  \\
& 50\% &  78.44 $\pm$ 0.74  &  77.02 $\pm$ 1.03  &   77.71 $\pm$  0.90 \\
& 75\% &  77.49 $\pm$ 0.66  &  78.59 $\pm$ 0.45  &  77.98 $\pm$ 0.62 \\\midrule

\end{tabular}}
\end{center}
\caption{Results for VGG-16. Reported values are mean and standard deviation over 3 iterations. *Sparsity is set to the closest smaller valid data sparsity, e.g. 1 channel is removed for $50\%$ sparsity.}
\label{table:vgg16}
\end{table}

\begin{table}[!htbp]
\begin{center}
\resizebox{0.45\textwidth}{!}{
\renewcommand{\arraystretch}{1.1}
\begin{tabular}{l|l||c c c}
\cmidrule[1pt]{3-5} 
\multicolumn{2}{c}{}&\multicolumn{1}{c}{Prune Weights }&\multicolumn{1}{c}{Prune Data*}&\multicolumn{1}{c}{Prune Both}\\\midrule
\multicolumn{2}{c}{\textbf{CIFAR10}} & \multicolumn{2}{c}{}             &  \\\midrule
\multirow{3}{4mm}{\begin{sideways}\parbox{12mm}{Sparsity}\end{sideways}}
& 25\% &  94.28 $\pm$ 0.13  & 94.52 $\pm$ 0.17  & 93.99 $\pm$ 0.13  \\
& 50\% &  93.89 $\pm$ 0.10  & 94.62 $\pm$ 0.10  &   92.53 $\pm$ 0.11   \\
& 75\% &  90.93 $\pm$ 0.09   &  93.08 $\pm$ 0.11  &  91.23 $\pm$ 0.19   \\\midrule

\multicolumn{2}{c}{\textbf{CIFAR100}} & \multicolumn{2}{c}{}             &  \\\midrule
\multirow{3}{4mm}{\begin{sideways}\parbox{12mm}{Sparsity}\end{sideways}}
& 25\% & 74.03 $\pm$ 0.69  &  76.96 $\pm$ 0.10  &  74.18 $\pm$ 0.23  \\
& 50\% &  65.63 $\pm$ 0.72  &  75.68 $\pm$ 0.05  &  72.14 $\pm$ 0.12   \\
& 75\% &  65.57 $\pm$ 0.40  &  70.17 $\pm$ 0.20  & 66.40  $\pm$ 0.10 \\\midrule

\multicolumn{2}{c}{\textbf{Caltech101}} & \multicolumn{2}{c}{}             &  \\\midrule
\multirow{3}{4mm}{\begin{sideways}\parbox{12mm}{Sparsity}\end{sideways}}
& 25\% &  82.26 $\pm$ 0.71 &  82.11 $\pm$ 0.80  &  82.18 $\pm$ 0.38  \\
& 50\% &  81.71 $\pm$ 0.80  &  83.08 $\pm$ 0.61  &  82.51 $\pm$ 0.13   \\
& 75\% &  80.09 $\pm$ 0.45  &  81.99  $\pm$ 0.49  & 80.50 $\pm$ 0.81 \\\midrule

\end{tabular}}
\end{center}
\caption{Results for ResNet18. Reported values are mean and standard deviation over 3 iterations. *Sparsity is set to the closest smaller valid data sparsity, e.g. 1 channel is removed for $50\%$ sparsity.}
\label{table:resnet18}
\end{table}

\begin{figure}[!htb]
\centering
 \includegraphics[width=0.53\textwidth]{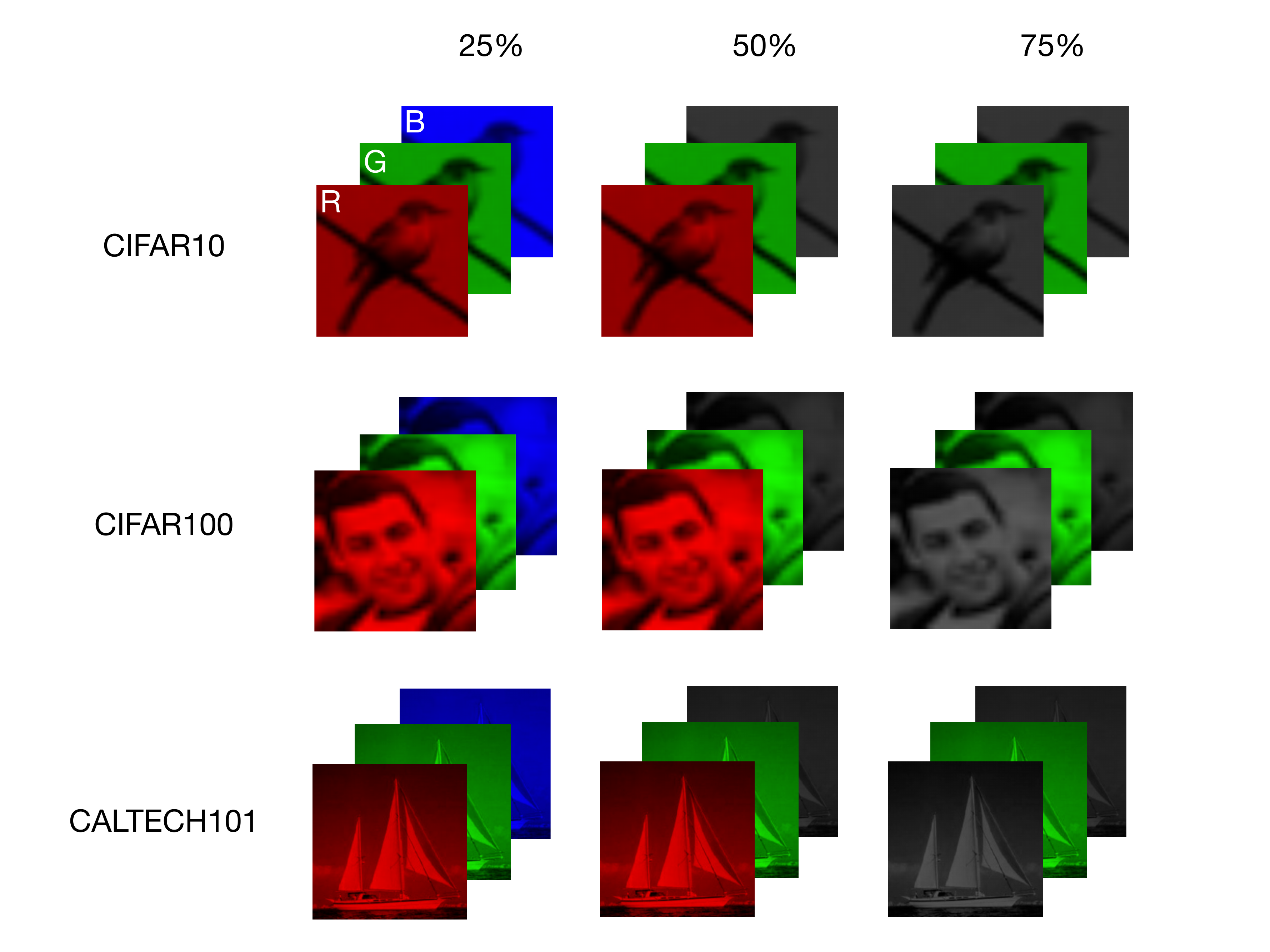}
\caption{Remaining channels after performing weight pruning. Channel-wise sparsity is set as the smaller closest valid sparsity to the target weight sparsity.}
 \label{mask_cnn}
\end{figure}

After the first step of our pruning experiment, we have 3 cases of data removal. We remove either 0, 1 or 2 channels from our input image. When 1 channel is removed, we find that our experiments consistently removes the blue channel. When 2 channels are removed, then the blue and red are the most often removed, as depicted in Figure \ref{mask_cnn}. Those observations correlates with how the human vision system works, i.e. more weight is given to green colors, then red and finally blue when mixing colors for grayscale conversion \cite{color}. 

We report in Table \ref{tab:budget_vgg} and Table \ref{tab:budget_rn} the parameters and FLOPs gains after applying each pruning method to $50\%$ sparsity. In the case of VGG16, it should be noted that fully-connected layers are responsible for more than $90\%$ of total parameters in the network, hence why the contribution of weight pruning is very small for the parameters gain. We observe substantial parameters and FLOPs gains when actually removing sparse structures from our network and their corresponding input channels.

\begin{table}[h!]
  \begin{center}
    \setlength\tabcolsep{4pt}
    \small
    \renewcommand{\arraystretch}{1.1}
    \begin{tabular}{l|cc}
    \cmidrule[1pt]{2-3}  
      & \multicolumn{1}{c}{Parameters (M)} & \multicolumn{1}{c}{FLOPs (G)} \\
    \hline
  {{\textbf{Baseline}}} & $134.32$ & $ 15.47 $ \\
  {{\textbf{Prune Weights}}} & $ 126.94 \ (\downarrow 5.49 \%) $ & $ 7.81  \ (\downarrow 49.51\%) $ \\
  {{\textbf{Prune Data}}} & $ 134.29  \ (\downarrow 0.22 \%)$ & $ 15.44  \ (\downarrow 0.19 \%) $ \\
      
  {{\textbf{Prune Both}}} & $ 71.88  \ (\downarrow 46.49\%)$ & $ 3.91  \ (\downarrow 74.72\%) $  \\
      \bottomrule
    \end{tabular}
  \end{center}    
    \caption{Parameter and FLOPs gains for VGG16. Pruning to $50\%$ weight sparsity and/or $33\%$ of data is applied.}
    \label{tab:budget_vgg}
\end{table}

\begin{table}[h!]
  \begin{center}
    \setlength\tabcolsep{4pt}
    \small
    \renewcommand{\arraystretch}{1.1}
    \begin{tabular}{l|cc}
    \cmidrule[1pt]{2-3}  
      & \multicolumn{1}{c}{Parameters (M)} & \multicolumn{1}{c}{FLOPs (M)} \\
    \hline
  {{\textbf{Baseline}}} & $11.18$ & $ 1813.57$ \\
  {{\textbf{Prune Weights}}} & $5.59 \ (\downarrow 50.00 \%)$ & $926.45 \ (\downarrow 48.92 \%)$ \\
      
  {{\textbf{Prune Data}}} & $11.17 \ (\downarrow 0.09 \%)$ & $ 1774.23 \ (\downarrow 2.17\%)$ \\
      
  {{\textbf{Prune Both}}} & $2.80 \ (\downarrow 74.95\%)$  & $463.28 \ (\downarrow 74.45\%)$  \\
      \bottomrule
    \end{tabular}
  \end{center}    
    \caption{Parameter and FLOPs gains for ResNet18. Pruning to weight $50\%$ sparsity and/or $33\%$ of data is applied.}
    \label{tab:budget_rn}
\end{table}

Besides having less parameters and operations, as the data pruning is applied in the same way for every input data, additional data storage gains can be obtained by only storing the relevant channels for our sparsity level, thus reducing storage costs by 1/3 or 2/3, depending if we keep 1 or 2 channels.

\section{\uppercase{Conclusions}}
\label{sec:conclusion}
Performing pruning on input data not only is interesting from a performance point of view, but also gives us valuable information about input data. Indeed, in the case of MLP, pruning less important pixels highlights where in the image is the most important information comprised. As illustrated in Figure \ref{mask_mlp}, those important pixels are generally comprised in the central zone of images, which can be explained by the bias of photographers usually centering important objects in an image \cite{photo_bias}. In the case of ConvNets, due to their inductive biases introduced by weight sharing, the information about individual pixels is lost as the same weights are applied to all pixels of an input image. However, we can still extract information about the most important channels in our input data. Removing those less useful portions of images allows to decrease the size and computations of our networks, but also can be used to lower the storage footprint of our dataset, as we have demonstrated that our networks were at least as efficient as when trained with complete images. \\

By following our 3-steps method, we also show that this is possible to go further than weight compression to maximize model speed-up. In particular, we show that pruning both data and weights allows to reach substantial gains in parameter counts and FLOPs. \\

Future work could involve investigating different and more recent vision architectures such as attention-based or patch-based architectures, which may provide different kind of feedback information about training data.

\bibliographystyle{apalike}
{\small
\bibliography{main}}







\end{document}